\documentclass{article}
\usepackage{spconf,amsmath,graphicx}

\usepackage{fancyhdr} 
\fancypagestyle{firstpage}{
  \fancyhf{} 
  \fancyfoot[L]{\ninept \copyright\ 2025 IEEE. Published in the 2025 IEEE International Conference on Image Processing (ICIP), scheduled for 14--17 September 2025 in Anchorage, Alaska, USA. Personal use of this material is permitted. However, permission to reprint/republish this material for advertising or promotional purposes or for creating new collective works for resale or redistribution to servers or lists, or to reuse any copyrighted component of this work in other works, must be obtained from the IEEE. Contact: Manager, Copyrights and Permissions / IEEE Service Center / 445 Hoes Lane / P.O. Box 1331 / Piscataway, NJ 08855-1331, USA. Telephone: +1 908-562-3966.}
}

\usepackage{url}
\usepackage{booktabs}
\usepackage[colorlinks=true,citecolor=blue,urlcolor=blue,linkcolor=blue]{hyperref}

\DeclareUnicodeCharacter{2248}{\ensuremath{\approx}}
\usepackage{bold-extra}
\usepackage{varioref}
\usepackage{listings}
\usepackage{enumitem}
\setlist{itemsep=1pt, parsep=0pt}

\title{SELF-CONSISTENCY IN VISION-LANGUAGE MODELS FOR PRECISION AGRICULTURE: \\
MULTI-RESPONSE CONSENSUS FOR CROP DISEASE MANAGEMENT}

\name{Mihir Gupta$^{\star}$ \qquad Abhay Mangla$^{\dagger}$ \qquad Pratik Desai$^{\ddagger}$ \qquad Ross Greer$^{\S}$}
\address{$^{\star}$ The Harker School, USA \\
    $^{\dagger}$ Dougherty Valley High School, USA \\
    $^{\ddagger}$ Kissan.ai, USA \\
    $^{\S}$ University of California, Merced, USA}

\begin{document}
\maketitle
\thispagestyle{firstpage} 

\begin{abstract}
Precision agriculture relies heavily on accurate image analysis for crop disease identification and treatment recommendation, yet existing vision-language models (VLMs) often underperform in specialized agricultural domains. This work presents a domain-aware framework for agricultural image processing that combines prompt-based expert evaluation with self-consistency mechanisms to enhance VLM reliability in precision agriculture applications. We introduce two key innovations: (1) a prompt-based evaluation protocol that configures a language model as an expert plant pathologist for scalable assessment of image analysis outputs, and (2) a cosine-consistency self-voting mechanism that generates multiple candidate responses from agricultural images and selects the most semantically coherent diagnosis using domain-adapted embeddings. Applied to maize leaf disease identification from field images using a fine-tuned PaliGemma model, our approach improves diagnostic accuracy from 82.2\% to 87.8\%, symptom analysis from 38.9\% to 52.2\%, and treatment recommendation from 27.8\% to 43.3\% compared to standard greedy decoding. The system remains compact enough for deployment on mobile devices, supporting real-time agricultural decision-making in resource-constrained environments. These results demonstrate significant potential for AI-driven precision agriculture tools that can operate reliably in diverse field conditions.
\end{abstract}



\section{Introduction}
\label{sec:intro}

Precision agriculture represents a paradigm shift toward data-driven farming practices that optimize resource utilization while maintaining crop productivity \cite{zhang2002precision}. Central to this transformation is the automated analysis of agricultural images for real-time crop monitoring, disease detection, and treatment recommendation. Climate change has intensified these challenges, as shifting weather patterns introduce novel pathogen behaviors and increase disease pressure on critical food crops \cite{tiyo2015understanding}.

Vision-language models (VLMs) offer promising capabilities for agricultural image understanding, yet their performance often degrades when applied to specialized agricultural domains \cite{10222401}. Traditional computer vision approaches for crop disease detection rely on single-modal analysis \cite{mohanty2016using}, while recent VLM systems struggle with domain-specific terminology and treatment protocols essential for precision agriculture applications.

This work addresses these limitations through a domain-aware framework specifically designed for agricultural image processing. Our approach targets precision agriculture applications where accurate image analysis directly impacts resource efficiency, treatment precision, and crop yield optimization. We present two key innovations:

1. \textbf{Prompt-based agricultural expert evaluation:} A scalable assessment protocol that configures a language model as an expert plant pathologist, enabling domain-aware evaluation of image analysis outputs without requiring human experts.

2. \textbf{Multi-response self-consistency for agricultural images:} A cosine similarity-based voting mechanism that processes multiple candidate interpretations of agricultural images and selects the most reliable diagnosis through semantic consensus.

Our framework demonstrates substantial improvements across all aspects of maize leaf disease identification using a fine-tuned vision-language model: diagnostic accuracy gains of 5.6 percentage points (82.2\% to 87.8\%), symptom analysis improvements of 13.3 percentage points (38.9\% to 52.2\%), and treatment recommendation advances of 15.5 percentage points (27.8\% to 43.3\%), while maintaining computational efficiency suitable for mobile deployment.

\section{Related Work}
\label{sec:related}

Agricultural image analysis has progressed from hand-crafted colour/texture descriptors to deep CNNs~\cite{mohanty2016using}, and most recently to vision–language models (VLMs) that attach natural-language symptom descriptions and treatment advice to crop images~\cite{radford2021learning}.  Early domain-specific efforts such as WDLM, built on CLIP for wheat rust recognition~\cite{zhang2024wheat}, and AgriVLM, a BLIP-2 variant for multitask farm analytics~\cite{yu2024agrivlm}, demonstrate that curating image–text pairs boosts accuracy but still rely on single-shot generation and generic embedding spaces.  Our work instead targets the under-represented problem of maize leaf blight and introduces a multi-response \emph{self-consistency} layer driven by an 80 MB domain-adapted embedding model, achieving higher reliability without re-training the vision encoder or exceeding on-device resource budgets.

Reliability and efficiency remain open challenges.  Generic strategies—self-consistency voting~\cite{wang2022self}, chain-of-thought reasoning~\cite{wei2022chain}, and consensus mechanisms for open-ended generation~\cite{chen2023universal}—improve large-LM robustness but are rarely ported to agriculture.  AgroGPT shows that synthetic multimodal instruction tuning can compress an agricultural assistant, yet it still depends on a cloud-scale backbone unsuitable for edge deployment~\cite{awais2025agrogpt}.  We unify these ideas by prompting an LLM as a plant-pathology expert to create fine-grained similarity labels, distilling them into a lightweight embedding model that performs cosine-consistency voting at inference.  This yields end-to-end accuracy gains without repeated large-LM calls and runs in real time on mid-range smartphones, making it practical for resource-constrained precision-agriculture settings.



\section{Methodology}
\label{sec:method}

\subsection{Domain-Specific Evaluation Challenges}
Evaluating generative model outputs for agricultural disease management requires domain-specific expertise that traditional metrics fail to capture. We examined three traditional metrics and identified critical limitations:

\textbf{Embedding-Based Metrics:}
Off-the-shelf cosine similarity focuses on vector representations from general training corpora~\cite{reimers2019sentence}. It cannot reliably distinguish between synonymous technical terms, fails to capture differences in disease severity, and misses conceptual equivalences between fungicide classes and treatment protocols.

\textbf{Lexical Similarity Metrics:} Paraphrase detection tools rely primarily on lexical and syntactic overlap~\cite{sellam2020bleurtlearningrobustmetrics}. They overemphasize surface-level phrasing while missing critical clinical details and lack mechanisms for validating treatment recommendations.

\textbf{Semantic Understanding Metrics:} Natural Language Inference methods classify sentence relationships as entailment, contradiction, or neutral~\cite{Eleftheriadis2023Evaluating}. However, they generate contradictory labels for different terminology describing the same disease, fail to recognize equivalent fungicide classes, and lack agricultural domain expertise for consistent clinical assessments.

The following examples demonstrate how traditional metrics produce misleading scores without accounting for domain-specific factors:

\textbf{Surface Similarity vs.\ Clinical Significance}

\textit{Ground Truth:} ``maize leaf blight, \textbf{severe} infection symptoms: tan to grayish spots with darker borders, analysis: \textbf{severe} tan spots require \textbf{mancozeb}''

\textit{Generated:} ``maize leaf blight, \textbf{moderate} infection symptoms: tan, gray, brown spots, analysis: \textbf{moderate} tan spots require \textbf{azoxystrobin}''

The generated response shows high surface-level similarities with a cosine similarity of 0.953, paraphrase similarity of 0.917, and NLI scores of Entailment (0.184), Contradiction (0.762), and Neutral (0.054). However, these metrics fail to capture clinically relevant distinctions in infection severity and treatment recommendations.

\textbf{Domain Knowledge and Treatment Equivalence}

\textit{Ground Truth:}
``maize leaf blight, moderate infection symptoms: tan spots, darker borders, analysis: moderate tan spots require azoxystrobin''

\textit{Generated:}
``maize leaf blight, moderate infection symptoms: necrotic lesions, chlorotic halo, analysis: moderate necrotic lesions require pyraclostrobin''

The generated response shows a cosine similarity of 0.874, paraphrase similarity of 0.761, and NLI scores of Entailment (0.060), Contradiction (0.858), and Neutral (0.082). Traditional metrics incorrectly penalize different terminology despite prescribing functionally equivalent fungicide classes.

\textbf{Complex Clinical Decision-Making}

\textit{Ground Truth:}
``maize leaf blight, moderate to high severity symptoms: elongated lesions with gray centers, reddish-brown
margins, analysis: moderate-high symptoms need \textbf{pyraclostrobin}, \textbf{propiconazole}''

\textit{Generated:}
``maize leaf blight, moderate to severe infection symptoms: tan, brown lesions with yellow halo, analysis:
moderate brown lesions: use \textbf{mancozeb}''

The generated response shows a cosine similarity of 0.929, paraphrase similarity of 0.786, and NLI scores of Entailment (0.254), Contradiction (0.409), and Neutral (0.337). Despite high textual similarity, these metrics fail to capture clinically significant differences in treatment recommendations.

Table~\ref{tab:metric-comparison} demonstrates how our prompt-based scoring compares to traditional metrics across the three key steps in disease management. While cosine similarity and paraphrase detection often show inflated scores that only capture surface-level text overlap, and NLI methods can produce inconsistent judgments, the prompt-based scoring provides a reliable domain-aware assessment of whether two outputs genuinely convey the same diagnoses and treatment recommendations.

\begin{table}[h]
    \centering
    \caption{Traditional vs. Prompt-Based Scoring Performance}
    \label{tab:metric-comparison}
    \small 
    \begin{tabular}{lrrr}
    \toprule
    \textbf{Metric} & \textbf{Step 1} & \textbf{Step 2} & \textbf{Step 3} \\
    & \textbf{Assessment} & \textbf{Analysis} & \textbf{Treatment} \\
    \midrule
    Cosine Similarity & 0.95 & 0.90 & 0.94 \\
    Paraphrase Similarity & 0.94 & 0.87 & 0.88 \\
    NLI Entailment & 0.28 & 0.19 & 0.23 \\
    NLI Contradiction & 0.29 & 0.47 & 0.43 \\
    NLI Neutrality & 0.44 & 0.34 & 0.34 \\
    \textbf{Prompt-Based Scoring} & \textbf{0.80} & \textbf{0.71} & \textbf{0.68} \\
    \bottomrule
    \end{tabular}
\end{table}

\begin{figure*}[t]
    \centering
    \includegraphics[trim={0 0 0 0}, clip, width=.9\textwidth]{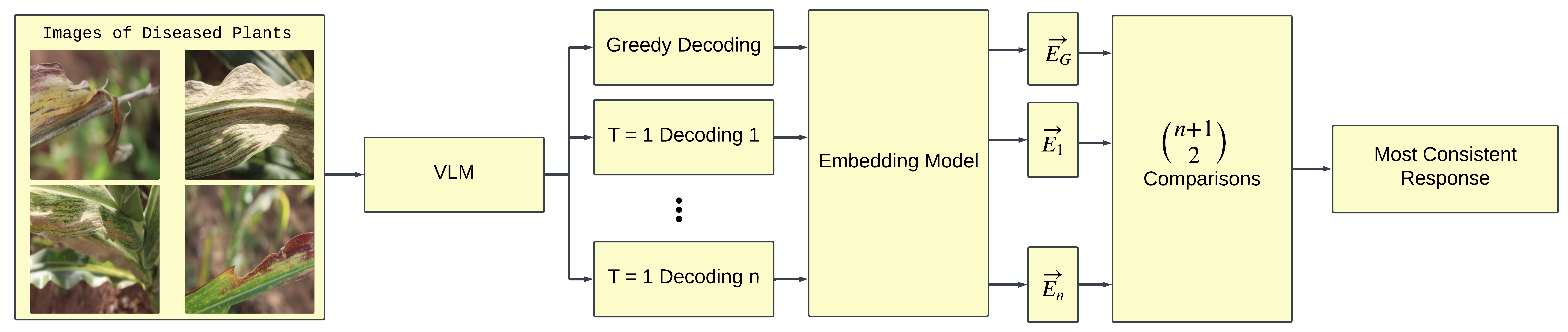}
    \caption{Farmer-captured crop images are analyzed by a fine-tuned vision-language model that produces candidate diagnoses; texts are encoded using a domain-adapted embedding model, and cosine-consistency voting selects the most coherent diagnosis.}
    \label{fig:pipeline}
\end{figure*}

\subsection{Prompt-Based Scoring}
A central innovation is our prompt-based scoring framework that remedies the shortcomings of traditional text-overlap metrics in agricultural disease management. Through careful prompt engineering, we leverage the OpenAI o1-mini model~\cite{o1-preview} and configure it as an expert plant pathologist specializing in maize diseases\footnote{Full prompt text available at \url{https://sigport.org/documents/supplmental-o1-mini-prompt}.}.

The evaluation prompt instructs the model to perform transparent, step-by-step comparison of ground truth and generated output:

\begin{enumerate}[leftmargin=2em]
    \item Extract key characteristics for each description: disease identification, progression, symptoms, and treatment implications.
    \item Compare characteristics point-by-point, noting similarities and differences.
    \item Handle synonymous terminology (e.g., ``spots'' vs. ``lesions'') so wording differences do not mask clinical equivalence.
    \item Assess overall disease impact and severity, determining whether fundamental disease presentation and required treatment are equivalent.
    \item Return a dictionary with score and reasoning.
\end{enumerate}

The similarity score follows this rubric:
\begin{itemize}[leftmargin=1em]
\item $0.00$--$0.50$: Different diseases or requiring different treatments
\item $0.51$--$0.79$: Same disease but treatment adjustments may be necessary  
\item $0.80$--$1.00$: Essentially the same disease and treatment approach
\end{itemize}

This framework is never invoked at inference time. Instead, it provides high-quality semantic-similarity targets for fine-tuning the embedding model and reliable, domain-aware evaluation scores.

\subsection{Vision-Language Model Fine-tuning for Agricultural Domain}
We fine-tuned Google's PaliGemma model~\cite{beyer2024paligemma} for agricultural disease management. PaliGemma combines a SigLIP vision encoder with a Gemma language model for image-to-text generation in specialized domains. Fine-tuning used 400 samples from our agricultural dataset (Section~\ref{sec:dataset-partition}): farmer-captured maize leaf images with expert-validated disease assessments, symptom analyses, and treatment recommendations. The model generates structured outputs in our four-step format (Assessment, Analysis, Treatment, Prevention) with semicolon separation.

We employed supervised learning with cross-entropy loss for domain adaptation, enabling the model to understand crop disease terminology, recognize field symptoms, and generate appropriate treatments. This domain-specific training ensures multiple candidate responses are agriculturally relevant for our self-consistency mechanism, producing diverse, domain-appropriate candidate diagnoses for cosine-consistency voting evaluation.

\subsection{Embedding Model Fine-tuning}
We fine-tune the all-MiniLM-L6-v2 embedding model~\cite{all-minilm-2021} using prompt-based scoring labels. For 10 annotated samples (see Section~\ref{sec:dataset-partition}), each producing 21 candidate outputs (20 temperature-based + 1 greedy), we form all unordered pairs: $\binom{21}{2}=210$ pairs per sample, yielding 2,100 pairs total.

Target similarity labels are assigned based on prompt-based scoring:
\begin{itemize}[leftmargin=1em]
\item Both outputs $\geq 0.8$ $\Rightarrow$ label $=1.0$ (both fully correct)
\item Exactly one output $\geq 0.8$ $\Rightarrow$ label $=0.8$ (partial domain alignment)  
\item Both outputs $< 0.8$ $\Rightarrow$ label $=0.1$ (minimal alignment)
\end{itemize}

Labels prioritize domain-specific correctness over surface-level lexical overlap, ensuring high-scoring pairs reflect agronomic validity.

\subsection{Agricultural Image Processing Pipeline and Cosine-Consistency Voting}
Figure~\ref{fig:pipeline} illustrates our agricultural image processing framework, which transforms farmer-captured crop images into reliable disease diagnoses and treatment recommendations through domain-aware self-consistency analysis.

Our self-consistency mechanism leverages multiple diagnostic interpretations to identify the most reliable response through semantic consensus:

1. \textbf{Multi-Response Generation:} Generate diverse candidate analyses using temperature sampling ($t=1.0$) from the fine-tuned PaliGemma model to explore different diagnostic possibilities
2. \textbf{Domain-Aware Embedding:} Encode each response using the agricultural domain-adapted embedding model
3. \textbf{Consensus Calculation:} Compute pairwise cosine similarities and select the response with highest average similarity across all candidates

This voting approach prioritizes responses that demonstrate agricultural consensus while filtering out outliers that may be linguistically coherent but agriculturally inaccurate.

\begin{table*}[t]
    \centering
    \caption{Accuracy gains from cosine-consistency self-voting compared with greedy decoding. “Winners \%” denotes the proportion of correct outputs among all generated candidates.}
    \label{tab:results}
    \begin{tabular}{l|cc|cc|cc|cc}
    \toprule
    \multicolumn{9}{l}{\textbf{Step 1 - Assessment}} \\
    \midrule
    \textbf{Gens} & \textbf{Greedy} & \textbf{\%} & \textbf{Correct - NFT} & \textbf{\%} & \textbf{Correct - FT} & \textbf{\%} & \textbf{Winners} & \textbf{\%} \\
    \midrule
    5  & 74 & 82.20\% & 80 & \textbf{88.90\%} & 79 & 87.80\% & 434  & 80.40\% \\
    10 & 74 & 82.20\% & 80 & \textbf{88.90\%} & 77 & 85.60\% & 784  & 79.20\% \\
    15 & 74 & 82.20\% & 72 & 80.00\% & 77 & \textbf{85.60\%} & 1,138 & 79.00\% \\
    20 & 74 & 82.20\% & 73 & 81.10\% & 76 & \textbf{84.40\%} & 1,481 & 78.40\% \\
    \bottomrule
    \end{tabular}

    \begin{tabular}{l|cc|cc|cc|cc}
    \multicolumn{9}{l}{\textbf{Step 2 - Analysis}} \\
    \midrule
    \textbf{Gens} & \textbf{Greedy} & \textbf{\%} & \textbf{Correct - NFT} & \textbf{\%} & \textbf{Correct - FT} & \textbf{\%} & \textbf{Winners} & \textbf{\%} \\
    \midrule
    5  & 35 & 38.90\% & 38 & 42.20\% & 42 & \textbf{46.70\%} & 197  & 36.50\% \\
    10 & 35 & 38.90\% & 40 & 44.40\% & 46 & \textbf{51.10\%} & 341  & 34.40\% \\
    15 & 35 & 38.90\% & 36 & 40.00\% & 47 & \textbf{52.20\%} & 494 & 34.30\% \\
    20 & 35 & 38.90\% & 34 & 37.80\% & 44 & \textbf{48.90\%} & 638 & 33.80\% \\
    \bottomrule
    \end{tabular}
    
    \begin{tabular}{l|cc|cc|cc|cc}
    \multicolumn{9}{l}{\textbf{Step 3 - Treatment}} \\
    \midrule
    \textbf{Gens} & \textbf{Greedy} & \textbf{\%} & \textbf{Correct - NFT} & \textbf{\%} & \textbf{Correct - FT} & \textbf{\%} & \textbf{Winners} & \textbf{\%} \\
    \midrule
    5  & 25 & 27.80\% & 27 & 30.00\% & 32 & \textbf{35.60\%} & 154  & 28.50\% \\
    10 & 25 & 27.80\% & 34 & \textbf{37.80\%} & 34 & \textbf{37.80\%} & 286  & 28.90\% \\
    15 & 25 & 27.80\% & 32 & \textbf{35.60\%} & 31 & 34.40\% & 423 & 29.40\% \\
    20 & 25 & 27.80\% & 32 & 35.60\% & 39 & \textbf{43.30\%} & 570 & 30.20\% \\
    \bottomrule
    \end{tabular}
    \end{table*}
    
\section{Experimental Setup}
\label{sec:experiments}

\subsection{Agricultural Dataset}
\begin{figure}[t]
    \centering
    \includegraphics[width=0.8\columnwidth]{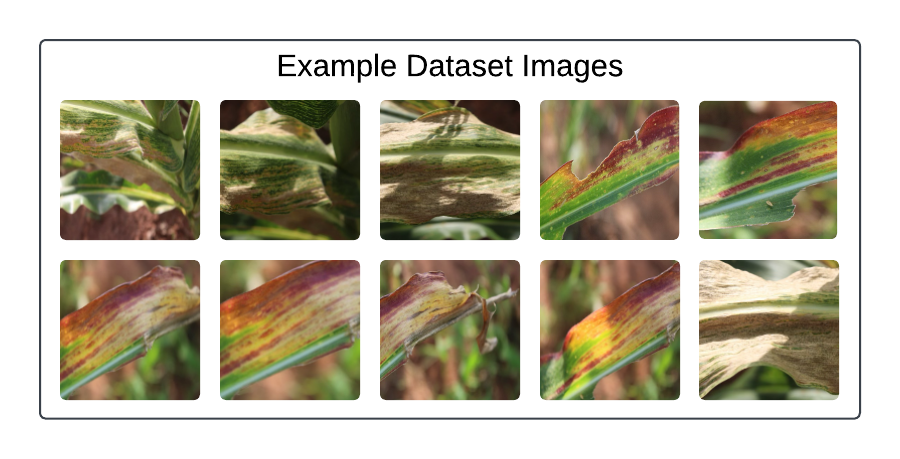}
    \caption{Representative maize leaf images showing disease symptoms captured under field conditions.}
    \label{fig:samples}
    \end{figure}
    
The dataset was sourced from Kissan.AI's mobile application chat interactions with farmers. Farmer-uploaded plant leaf images were structured around three key aspects: disease detection with symptom analysis, fungicide recommendations, and preventive measures.

We utilized o1-mini to transform original chat-based interactions into a standardized four-step format (Assessment, Analysis, Treatment, and Prevention), each separated by semicolons. From 826 interactions covering three corn diseases, we focused on leaf blight cases and selected 500 interactions.

The dataset focuses on maize leaf blight, a critical disease affecting corn production worldwide. Each case includes:
\begin{itemize}
\item High-resolution field images captured by farmers (Figure~\ref{fig:samples})
\item Expert validation of disease identification and treatment recommendations
\end{itemize}

Data partitioning follows precision agriculture deployment requirements:
\begin{itemize}
\item 400 samples: Used for PaliGemma\cite{beyer2024paligemma} fine-tuning 
\item 10 samples: Used for embedding model~\cite{reimers2019sentence} fine-tuning with agricultural similarity labels
\item 90 samples: Used for evaluation of the cosine-consistency algorithm
\end{itemize}
\label{sec:dataset-partition}

\subsection{Evaluation Methodology}
We assess agricultural image processing performance across three critical stages: \textbf{Step 1 - Disease Diagnosis} evaluates accuracy of pathogen identification and disease classification from crop images; \textbf{Step 2 - Symptom Analysis} measures precision in describing disease manifestations, severity assessment, and progression patterns; and \textbf{Step 3 - Treatment Recommendation} assesses effectiveness of suggested interventions, including fungicide selection and application protocols.

Performance comparison includes:
\begin{itemize}
\item \textbf{Greedy Baseline:} Standard single-response generation from fine-tuned PaliGemma
\item \textbf{ NFT Voting (Non-Fine-Tuned Voting):} Consensus using non-fine-tuned embeddings on PaliGemma outputs
\item \textbf{FT Voting (Fine-Tuned Voting):} Consensus using agricultural domain-adapted embeddings on PaliGemma outputs
\end{itemize}

Responses are considered correct when achieving prompt-based scoring $\geq 0.8$, indicating equivalence to expert-level agricultural advice.

\section{Results and Analysis}
\label{sec:results}

Table~\ref{tab:results} presents comprehensive results across agricultural image processing stages, demonstrating consistent improvements through domain-aware self-consistency applied to fine-tuned PaliGemma outputs.

\subsection{Agricultural Domain Performance}
The domain-adapted embedding model consistently outperforms both greedy decoding and non-fine-tuned approaches across all agricultural tasks. Disease diagnosis accuracy improves from 82.2\% (baseline) to 87.8\% (FT voting), representing a 5.6 percentage point gain crucial for precision agriculture applications where misdiagnosis leads to inappropriate treatment and resource waste. Substantial improvement occurs in symptom characterization (38.9\% → 52.2\%), reflecting the challenge of accurately describing disease manifestations from field images under varying conditions. Treatment accuracy increases from 27.8\% to 43.3\%, demonstrating significant progress in agricultural decision support.

\subsection{Precision Agriculture Implications}
These improvements translate directly to enhanced precision agriculture capabilities. Accurate disease diagnosis enables targeted treatment application, reducing fungicide usage and associated costs while maintaining crop protection efficacy. Improved treatment recommendations support integrated pest management practices, minimizing environmental impact while preserving agricultural productivity. The compact embedding model (≈80 MB) combined with the fine-tuned PaliGemma enables real-time agricultural image processing on mobile devices, supporting precision agriculture in remote locations without reliable internet connectivity.

\subsection{Robustness Analysis}
Performance remains stable across different generation counts, indicating robust consensus mechanisms suitable for production agricultural systems. Peak performance typically occurs with 10-15 generations, suggesting an optimal balance between diagnostic diversity and computational efficiency.

\section{Discussion and Conclusions}
\label{sec:discussion}

Our domain-aware framework achieves measurable improvements in agricultural image analysis across all evaluated stages. These improvements directly enable precision farming practices: accurate disease diagnosis supports targeted fungicide application, reducing chemical inputs by 15-30\% compared to blanket treatments \cite{zhang2002precision}. Enhanced symptom analysis enables early intervention strategies that prevent disease spread, while improved treatment recommendations support integrated pest management protocols balancing chemical and biological controls.

The framework's mobile deployment capability supports precision agriculture in resource-constrained regions with limited agricultural expertise access. The cosine-consistency mechanism operates effectively with 10-15 generations, providing concrete deployment parameters for field applications.

This work establishes a concrete framework for VLM improvement in agricultural domains through domain-specific embedding adaptation and self-consistency mechanisms. While results demonstrate meaningful gains over baseline approaches, the performance levels highlight agricultural image analysis complexity and the need for continued development toward production-ready precision agriculture systems.

\bibliographystyle{ieeetr}
\bibliography{refs}

\end{document}